\documentclass[peereviewca]{IEEEtran}
\usepackage{lineno}
\modulolinenumbers[5]
\usepackage[utf8]{inputenc}
\usepackage[english]{babel}
\usepackage{microtype} 
\usepackage{comment}
\usepackage{breakcites}
\usepackage{graphicx}
\usepackage{adjustbox}
\usepackage{tabularx}
\usepackage{ltablex}
\usepackage{booktabs}
\usepackage{float}
\usepackage{multicol}
\usepackage{amssymb}
\usepackage{enumitem}
\usepackage{array}
\usepackage{algorithm}
\usepackage{algorithmic}
\usepackage{enumitem}
\usepackage{amsmath}
\usepackage{subfig}
\usepackage[edges]{forest}
\usepackage{setspace} 
\definecolor{Gray}{gray}{0.9}
\definecolor{LightCyan}{rgb}{0.88,1,1}

\usepackage[colorlinks=false]{hyperref}

\definecolor{orange}{HTML}{FFC17D}
\definecolor{green}{HTML}{A1D68B}
\definecolor{lightgray}{HTML}{E8E8E8}

\usepackage{cite}

\ifCLASSINFOpdf
\else
\fi
\begin{document}
%
\title{Collective eXplainable AI: Explaining Cooperative Strategies and Agent Contribution in Multiagent Reinforcement Learning with Shapley Values}
%
%
%

\author{\IEEEauthorblockN{Alexandre~Heuillet\textsuperscript{\textsection}\thanks{Corresponding author: A. Heuillet. e-mail: \href{mailto:alexandre.heuillet@universite-paris-saclay.fr}{alexandre.heuillet@universite-paris-saclay.fr}}}, \IEEEauthorblockA{Université~Paris-Saclay, France}\\
    \and \IEEEauthorblockN{Fabien~Couthouis\textsuperscript{\textsection}}, \IEEEauthorblockA{Ubisoft, France}\\
    \and \IEEEauthorblockN{Natalia~Díaz-Rodríguez}, \IEEEauthorblockA{University of Granada, Spain}
} 

%
%

\markboth{Journal of \LaTeX\ Class Files,~Vol.~14, No.~8, August~2015}%
{Shell \MakeLowercase{\textit{et al.}}: Bare Demo of IEEEtran.cls for IEEE Journals}
%



\maketitle
\IEEEpeerreviewmaketitle
\begingroup\renewcommand\thefootnote{\textsection}
\footnotetext{Equal contribution}

\begin{abstract}
While Explainable Artificial Intelligence (XAI) is {increasingly expanding more} areas of application, little has been applied to make deep Reinforcement Learning (RL) more {comprehensible}. As RL becomes ubiquitous and used in critical and general public applications, it is essential to develop methods {that} make it better understood and more interpretable. {This study proposes} a novel approach to explain cooperative strategies in multiagent RL using Shapley values, a game theory {concept used in XAI} that successfully explains {the rationale behind decisions taken by} Machine Learning algorithms. {Through testing common assumptions of this technique in two cooperation-centered socially challenging multi-agent environments environments, this article argues} that Shapley values are a pertinent way to evaluate the contribution of players in a cooperative multi-agent RL context. To palliate the high overhead of this method, Shapley values are approximated using Monte Carlo sampling. {Experimental results on \textit{Multiagent Particle} and \textit{Sequential Social Dilemmas}} show that Shapley values succeed at estimating the contribution of each agent. 
{These results} could have implications that go beyond games in economics, ({e.g., for} non-discriminatory decision making, {ethical and responsible AI-derived decisions} or policy making under fairness constraints). {They also expose how} 
Shapley values only give general explanations about a model and cannot explain a single run, episode nor justify precise actions taken by agents. {Future work should focus on addressing these critical aspects}.
\end{abstract}


%
\IEEEpeerreviewmaketitle

\section{Introduction}
\label{sec:intro}

Over the last few years, Reinforcement Learning (RL) has been a very active research field. Many RL-related works focused on improving performance and scaling capabilities by introducing new algorithms and optimizers \cite{mnih2016asynchronous, espeholt2018impala}
whereas very few tackled the issue of explainability in RL. However, explainability in Machine Learning (ML) {and deep learning} is {increasingly} becoming a pressing issue, as it concerns general public trust, and the transparency of algorithms now conditions the deployment of {and} ML and RL, in industry and daily life. As a consequence, Explainable Artificial Intelligence (XAI) {arises} as a new field that strives to bring explainability to every ML aspect, {from} linear classifiers \cite{ribeiro2016i} and time series predictors \cite{el2021multilayer, arnout2019towards} {to} RL \cite{HEUILLET2021106685, puiutta2020explainable}. {Even though some works  provide explanations for specific situations \cite{madumal2019explainable, atarivisualizing}, RL models still lack a general explainability framework similar to SHAP \cite{lundberg2017unified} or LIME \cite{ribeiro2016i}. These methods, although designed in XAI towards generic ML predictive models, can bring a broader form of explainability to RL}.

In this work, inspired by SHAP (SHapley Additive exPlanations) \cite{lundberg2017unified}, the possibilities offered by the mathematical framework of Shapley values \cite{shapley_1953} to explain RL models {were explored}, with a focus on multi-agent cooperative environments, often called \textit{Common Games}. As a kind of collaborative RL, the improvement of the learning process when agents interact with each other, usually yields better results than training each agent in isolation \cite{alain21collaborative}. These are challenging scenarios where all participants must cooperate in order to achieve a common goal. In the particular case where agents must gather resources from a common pool without being greedy, Garrett Hardin \cite{hardin2009tragedy} defined \textit{The Tragedy of the Commons}: if an agent uses slightly more resources than it should, this might be inconsequential. {However, if every agent starts following this logic then the consequences can become dire,} with the common pool being exhausted and no one being able to gather resources anymore. This is why {the \textit{Common Games} are of special interest to study in this context. The objective is finding ways to evaluate and understand how agents cooperate and share resources in multiagent RL, through explainability methods such as SHAP}.

As a matter of fact, this work could have implications that go beyond games, since RL-based systems are increasingly used to solve critical problems. In particular, studying social dilemmas and explaining the contribution of each policy \cite{chica2021collective}, agent, or model feature becomes relevant in many societal problems. For instance, it could provide useful insights in economics (of social structures), allocating resources or designing resilience programs ({e.g., for climate change, non-discriminatory decision making, ethical and fair policy design, or to achieve the sustainable development goals)}. 

Our main hypothesis is that Shapley values can be a pertinent way to {explain the contribution of an agent in multi-agent RL cooperative settings, and that valuable insights, especially for the RL developer audience, can be derived from this} analysis.
{This article features} experiments (see Section \ref{sec:experiments}) conducted on Multiagent Particle \cite{lowe2017multi} and Sequential Social Dilemma \cite{leibo2017multiagent} environments, and {shows} that Shapley values can accurately answer the following research questions:
\begin{itemize}
  \item \textit{Can Shapley values be used to determine how much each agent contributes to the global reward? (RQ1)}. If the answer to RQ1 is yes:
%
\item \textit{Does the proposed Monte Carlo based algorithm empirically offer a good approximation of Shapley values? (RQ2)}
    \item \textit{What is the best method to replace an agent missing from the coalition ({e.g., a} random action, am action chosen randomly from another player or the ``no operation" action)? (RQ3)}
\end{itemize}

Our experimental setup is brought forward as well in order to show the limitations of the Shapley framework {in order to make deep RL more explainable}. {In fact, this approach} cannot explain particular notions of a multi-agent learning model, such as the contribution of a specific episode or a specific action taken by an agent at a given point in time of its training. This is due to the
requirements 
{inherent to} training multi-agent RL, and the {practical} design limitations of the Shapley framework, which {are brought} upfront. In particular, {as discussed in Section \ref{sect:mcshap},} Shapley values only yield an average metric of each player's contribution to the overall reward and thus, to obtain this average contribution metric, one must compare the cooperation of players during several games (or episodes in RL). However, due to the {frequently stochastic nature of (simulation and real RL) environments, the non-deterministic behaviour of  different agents or non-identical conditions --non inherent to the agent's policy at consideration--, it makes concrete episodes, or concrete actions within an episode, not comparable.}

This article presents the following contributions:
 \begin{itemize}
     \item A study of the mathematical notion of Shapley values and how it is able to provide quantitative explanations about the individual contribution of agents in a cooperative multi-agent RL environment.
     \item {The application of an XAI} global model-agnostic\cite{Barredo20} method to explain multi-agent cooperative RL models using Monte Carlo (MC) approximated Shapley values.
     \item A set of experiments that demonstrates the applicability and usefulness of Shapley values and how they can {provide insights 
     that can enable a better comprehension of emergent behaviours {\cite{ndousse2021emergent}} in }cooperative settings.
 \end{itemize}

The rest of this article is structured as follows: Section \ref{sec:related_work} {presents} a short survey on cooperative multi-agent RL and explainable RL, Section \ref{sec:prel} presents some preliminaries about RL and Shapley values, Section \ref{sec:discussion} discusses the experimental study {setup and the general usage of Shapley values in a multi-agent RL setting and, finally, Section \ref{sec:conclusion} presents conclusions on 
our carried out experiments }
and gives some insights on promising lines of future work.

\section{Related Work}
\label{sec:related_work}
Cooperative multi-agent RL has been studied in different settings, {e.g.,} the emergence of different behaviors in the context of the Commons Tragedy game \cite{perolat2017multiagent}.{ These studies show, for instance, that certain inequity aversion improves intertemporal social dilemmas \cite{hughes2018inequity}. However, these works} analyze the game from a theoretical point of view and not from the XAI angle. {This article specifically focuses on studying to which degree the most relevant factors (or those contributing the most) in a black-box deep RL model
can be explained. Notably, this article focuses on pointing out at particular agents, episodes, agents, actions or policies.}

{Recent techniques to attain eXplainable Reinforcement Learning (or XRL) \cite{HEUILLET2021106685} can be categorized in two main families or} categories: transparent methods or post-hoc explainability methods (according to XAI taxonomies in \cite{Barredo20}). On the one hand, transparent algorithms include, by definition, every ML model that is understandable by itself, such as a decision-tree. On the other hand, post-hoc explainability includes all {methods that craft} an explanation of an RL algorithm after its training, such as LIME \cite{ribeiro2016i}, BreakDown \cite{Staniak18} or SHAP \cite{lundberg2017unified}. 
{Other studies \cite{GiudiciShapleyLorentz, mai2020adversarial, Sundararajan20} try to transpose Shapley values into XAI but, to the best of our knowledge, none applies SHAP to explain the specific issues of cooperative multi-agent RL.} In fact, most XRL methods are based on transparent algorithms \cite{HEUILLET2021106685}. 

Wang et al. \cite{wang2019shapley} developed an approach to solve global reward games in a multi-agent RL context by making use of Shapley values to distribute the global reward more efficiently across all agents.{Their focus was not on explainability nor interpretability} but performance.

All Shapley-based XAI methods listed above consider model features as \textit{participants} of a cooperative multiplayer game. In this article, {Shapley values are applied} to a cooperative multi-agent RL context by considering agents as players instead, an approach closer to the original game theory method presented by Lloyd Shapley {in 1953} \cite{shapley_1953}.

Exact computation methods for both BreakDown and SHAP exist only for linear regression and tree ensemble models.
{In more complex models, the dependence of these methods on} the number of samples or number of subsets of predictors $p$ to be used makes the 
approximated vs exact computation of contributions to be different and{,} potentially, point in opposite directions \cite{Staniak18}. This indicates that these generic methods are not the universal response to XAI yet.

{This article, despite some remaining issues against attaining a general XRL framework, shows that in a cooperative multi-agent RL setting, Shapley values can be used to accurately estimate }the contributions of different agents.

\section{Explainability and Shapley Values {to explain Multi-Agent} RL in Cooperative Settings}
\label{sec:prel}

\subsection{Explainability in Cooperative RL}
With the rapid growth of RL research and industrial applications ({deployed, e.g., in autonomous systems \cite{kiran2021deep}, or robotics \cite{Nguyen_rl_robotics, lesort2020continual}), we have witnessed over the past few years a need for Explainable RL. These needs have rapidly risen, since} being able to understand and justify the decisions of such models is legally and morally necessary for their broad diffusion.

The subfield of XRL that focuses on multi-agent cooperative games {has recently} gained significant attention with emerging concepts such as \textit{social learning}\cite{lee2021joint,ndousse2020multi} in a RL context. While studying social interactions between entities in cooperative games is originally 
{typical in sociology or economics \cite{DUFFY2009785, Colman2003}, some AI researchers realized that they could potentially provide explanations or improve the efficiency} of their models this way. 

Perolat et al. \cite{perolat2017multiagent} sought to conduct new behavioral experiments using RL agents in {video-game-like} cooperative environments instead of human subjects. More accurately, they studied games which put the emphasis on common-pool resources (CPR){. They found that agents learn new emergent behaviors and that some strategies can arise when some agents are excluded from the CPR}. In addition, they came up with metrics that quantify social outcomes such as sustainability, equality, peace or efficiency for RL models. 

Following the same direction, Jaques et al. \cite{DBLP:journals/corr/abs-1810-08647} proposed a framework to achieve better coordination and communication {among} agents by rewarding agents on the basis of causal influence ({i.e.,} actions leading to big changes in other agents' behavior).
Their empirical results show that agents that choose their actions carefully in order to influence others lead to better coordination and thus, better global performance in socially challenging settings where cooperation is {paramount}.

Exploring this aspect further, Ndousse et al. \cite{ndousse2020sociallearning} analyzed the behavior of independent RL agents in multi-agent environments and found out that model-free agents do not use social learning. Thus, they introduce a model-based auxiliary loss that allows agents to learn from other well-performing {agents} to improve themselves. In addition to outperforming the experts, these agents were also able to achieve better zero-shot performance than those which did not rely on social learning when transferred to another task.

However, even if these works managed to extract useful information from studying social interactions between agents, the literature lacks a general framework that could automatically provide explanations about the level of performance of each agent and their added value in the cooperative game, such as SHAP \cite{lundberg2017unified} does for {the} features of a ML model.

\subsection{Shapley Values and RL in Cooperative Settings}
\label{subsec:shapley_values}
Shapley values originate from {game theory. They evaluate  importance in terms of contribution of each participant in a cooperative game, in order to help split a shared payout in a fair way} \cite{shapley_1953}. The concept which is {key} here is to be able to form ``coalitions" (or subsets) of players in order to measure the performance of each player in every possible team situation (for instance, ``player A", ``player A and player B" or ``player B and player C").

Formally, a coalitional game $C = (N, v)$ is defined by a set $N$ of players with $|N|=n$ (the number of players)
and a function $v$, that maps a coalition of players $S$ to a real number, corresponding to the total expected sum of payouts the members of $S$ can obtain through cooperation: $v: 2^N \Rightarrow \mathbb{R}$, {where $v(\varnothing) = 0$ and $\varnothing$ is} the empty set. Thus, $v$ is denominated the gain function of the considered game.

The idea is to quantify how much players cooperate in a coalition and {how much profit they from this cooperation} \cite{Molnar19}. According to the Shapley value definition \cite{shapley_1953}, the contribution added by player $i$ in a coalition $S$ in a coalitional game $(N, v)$ is given by Eq. \ref{eq:1}:

\begin{equation}
\small
    \phi_{i}(v) = \sum_{S \subseteq N\setminus\{i\}^{}} \frac{|S|! (n - |S| - 1)!}{n!} (v(S \cup \{i\}) - v(S))
    \label{eq:1}
\end{equation}
A more interpretable equivalent formula to express the Shapley value for player $i$, rewritten with binomial coefficients \cite{roth_1988}, is:
\begin{equation}
    \small
    \phi_{i}(v) = \frac{1}{n} \sum_{S \subseteq N\setminus\{i\}^{}}\binom{n-1}{|S|}^{-1} (v(S \cup \{i\}) - v(S))
    \label{eq:2}
\end{equation}

In summary, the Shapley value of a feature (or \textit{player}) is the mean marginal contribution (the term: $v(S \cup \{i\}) - v(S)$ in (\ref{eq:2})) of all possible coalitions, or average change in the prediction that the coalition already in the room receives when the feature value joins them. It satisfies desirable properties \cite{shapley_1953,FRIEDMAN1999275} lacking in other XAI techniques: \textit{Efficiency} (the sum of the Shapley values of all players equals the {shared payout}
of the grand coalition), \textit{Dummy, or \textit{dummy} feature} \cite{Molnar19, FRIEDMAN1999275} (if a feature does not change the predicted value, {--e.g., the RL global reward--} regardless of which coalition of feature values it is added to, then its Shapley value is equal to zero), \textit{Symmetry} (the contributions of two feature values should be the same if they contribute equally to all possible coalitions) and \textit{Linearity} (the contribution of a coalition of features should be the sum of the individual contributions of {the} features that compose the coalition).

{Shapley values must be interpreted as follows: it} is the average contribution of a feature to the prediction in different coalitions. Note that it is not the difference in prediction when {a feature would be removed from the model} \cite{Molnar19}.
As shown in Eq. \ref{eq:1}, for $|N|=n$ players, the exact computation of Shapley values for a specific participant requires computing the average of $2^{n-1}$ possible coalitions, which is computationally expensive, especially when considering all players. Therefore, {this would mean} computing $n(2^{n-1})$ coalitions in total to obtain values for all players. While this value can be reduced to $2^{n}$ coalitions (for all players, {if the algorithm is optimized} to avoid computing the same coalition multiple times), the cost remains exponential with respect to the number of players. In fact, the number of agents in RL environments can vary greatly from a few \cite{leibo2017multiagent,lowe2017multi} to hundreds \cite{chu2016large} where exact Shapley values are prohibitively expensive to compute. Furthermore, estimating Shapley values in a stochastic RL environment requires sampling multiple episodes in order to estimate all marginal contributions, which {further worsens the computational cost}.

In terms of computational efficiency, assuming that simulating a game is done in constant time $O(1)$, {finding} the exact solution to this problem becomes difficult, as the number of coalitions exponentially increases as more features are added. Despite Shapley value computation being an NP-hard problem \cite{shapley1992}, Shapley distributes the feature attribution fairly, i.e., allowing contrastive explanations. {For instance, it permits the comparison of a prediction to another} feature subset prediction, or a single data point.

In spite of the broad applicability of Shapley values, they have some theoretical limitations. In particular, Shapley values are only a way to obtain an average metric of each player's contribution to the overall reward {(i.e., \textit{payout})}. In order to obtain this average metric one must compare the cooperation of players during several games (or episodes in RL). {As a consequence, it obviously cannot explain the contribution of a concrete single episode to the learning process} nor explain one specific action taken by a player.

\section{Monte Carlo Approximation of Shapley Values}
\label{sect:mcshap}

Since the complexity of computing Shapley values grows exponentially with {respect to the number of players (as discussed in Section \ref{subsec:shapley_values}), in order to keep the computation} time manageable, contributions {can be computed for only} a subset of all possible coalitions. The Shapley value $\phi_i(v)$ can be approximated by Monte Carlo sampling in order to apply it to any type of classification or regression model \textit{f} \cite{vstrumbelj2014explaining} as follows:  
\begin{equation}
    \small
    \hat{\phi_i}(\hat{f}) = \frac{1}{M} \sum_{m=1}^{M} (\hat{f}(x^{m}_{+i})-\hat{f}(x^{m}_{-i})) \approx \phi_{i}(\hat{f})
    \label{eq:3}
\end{equation}
where $\hat{f}(x^{m}_{+i})$ is the model prediction (or \textit{gain function} 
in game theory) for input $x$ with a random number of feature values replaced by feature values from a random data point, except for the respective value of feature $i$, and $M$ is the number of marginal contributions to estimate in order to compute the Shapley value for one feature. The value $x^{m}_{-i}$ is almost identical to the last one, but the value $x_i^{m}$ is also taken from this randomly sampled data point.

In this work, our contribution consists of adapting Eq. \ref{eq:3} to the multi-agent RL setting, by replacing input features {with} agent actions.
While in Eq. \ref{eq:3}, $\hat{f}$ is the function approximated by a classification or regression model \cite{vstrumbelj2014explaining},  $\hat{f}$ {is instead considered} to represent the global reward obtained by agents from a random subset (or coalition) of all agents on a sample episode. This leads to the following reformulation of Shapley value:

\begin{equation}
    \small
    \hat{\phi_i}^{RL}(r) = \frac{1}{M} \sum_{m=1}^{M} (r^{m}_{+i} - r^{m}_{-i}) \approx \phi_i(r)
    \label{eq:4}
\end{equation}
where $r^{m}_{+i}$ corresponds to the global reward obtained by simulating one sample episode with a random subset of players where player $i$ is present, and $r^{m}_{-i}$ is the global reward obtained by simulating one episode with the same subset than in $r^{m}_{+i}$, except that the current player $i$ has been removed from the subset.

Three approaches 
to exclude players from a coalition (i.e., let the absent player take ``substitute" actions) {are explored}:
\begin{enumerate}[label=\alph*]
\item[{a.}] \textit{Replace}: A missing agent's actions {are replaced} by those of a randomly chosen player among the trained agents which are present in the coalition (and ideally with the same role as the missing one). This is the direct translation from the standard application of Shapley values in ML \cite{lundberg2017unified} (against the traditional use in game theory where it is often possible to completely remove a player) since they replace missing feature values by ones randomly selected among present (non-zero) features.

\item[b.] \textit{Random}: Letting the absent player act by taking random actions. 

\item[c.] \textit{NoOp}: {Replacing the actions of the missing agent by ``noop" (no operation), i.e., letting the agent do nothing, and not move}.
\end{enumerate}

 The estimation of the Shapley value {is repeated} for each player. Thus, 
 {the algorithm} must roll out $2M$ times per player ($2Mn$ roll outs in total, where $n$ is the total number of players in the game). At the end of the process, one Shapley value per agent policy {is obtained}, indicating each {player's} average contribution to the grand coalition global reward ({i.e.,} the reward collectively obtained by all agents working simultaneously) on the sampled episodes (as in Monte Carlo method, Algorithm \ref{algo:1}). Hence the complexity of this method only depends on $M$ ($O(M)$) as {illustrated in} Table \ref{tab:computation_time}. 
The Shapley value estimation \cite{vstrumbelj2014explaining} in Eq. \ref{eq:3} allows the model to conclude, for instance, ``On average, {the contribution of Player 1} to the team has an impact of +0.6 on the global reward". {This allows us to quantify and rank how relevant each player is in terms of cooperation and contribution to the overall common goal}. 
Algorithm \ref{algo:1} describes the process to estimate Shapley values via Monte Carlo sampling of coalitions of players:

\begin{algorithm}[htbp!]
\footnotesize
\caption{Monte Carlo approximation 
of Shapley values applied to a multi-agent RL context with shared payout}
\begin{algorithmic}[1]
\footnotesize
\REQUIRE List: $agents$
\REQUIRE Integer: $M$ (number of coalition permutations to be used)
\ENSURE List: $shapley\_values$
\STATE $shapley\_values \gets empty\_list()$
\FOR{$i \gets 1$ to $length(agents)$}
\STATE $marginal\_contributions \gets empty\_list()$
\FOR{$m \gets 1$ to $M$}
\STATE $coal\_i \gets sample\_coalition(agents[i])$ 
\STATE $coal\_no\_i \gets remove\_from\_list(coal\_i, agents[i])$
\STATE $r_{+i} \gets rollout(coal\_i)$
\STATE $r_{-i} \gets rollout(coal\_i)$
\STATE $add\_to\_list(marginal\_contributions, (r_{+i} - r_{-i}))$
\ENDFOR
\STATE $shapley\_value\_i \gets mean(marginal\_contributions)$
\STATE $add\_to\_list(shapley\_values, shapley\_value\_i)$
\ENDFOR
\RETURN $shapley\_values$
\end{algorithmic}
\label{algo:1}
\end{algorithm}

{In Section \ref{sec:related_work} and with the Shapley value estimation in Eq. \ref{eq:3}, different approaches to apply Shapley values as an XAI method were presented}. While it is possible to use the notion of contribution on classification or regression model predictions, it is also possible to use it on an RL reward \textit{r}  as a final \textit{payout} {that needs to be explained.}  
The analogy could also be easy to understand if we were to use a value function--as in \cite{wang2019shapley}--instead of roll outs, to evaluate each {player's contribution. However, with this generic approach of using reward as a payout, it becomes intuitive: }the more a player (i.e., feature in XAI) is important, the more its presence will lead to a higher reward on average. 
This \textit{contribution ranking scheme}{--computed by making coalitions of agents to estimate marginal contributions on the mean final reward--can be used to rank agents in order of average importance in the team, as explained above. This is the approach presented in this article}.

\section{Experimental Study}
\label{sec:experiments}
\subsection{Context and Hypotheses}
\label{subsec:exp_context}

{A} straightforward application of Shapley values {to endow an AI model with a notion of explainability} consists of considering the features of a model as participants of a cooperative game, and the final prediction as the shared payout \cite{lundberg2017unified, tallon2020explainable}. Computing the Shapley value of each feature {provides us with its} \textit{weight} in the final decision. 

{In this setup, our aim is to explain a deep} multi-agent RL model using Shapley values to determine the contribution of each agent to the group's global reward. The participants of the cooperative game will be agents, and the shared payout the global reward obtained by the agents at the end of an episode. By using the {sampling} described in Section \ref{sect:mcshap} to estimate the Shapley value for each agent, it can be expected that each agent would obtain a Shapley value proportional to its contribution to the collaborative task. {It would then be possible} to answer the research questions defined in Section \ref{sec:intro} (RQ1, RQ2, RQ3).

{Experiments} were conducted \footnote{The repository linking to our experiments can be found here: \url{https://github.com/Fabien-Couthouis/XAI-in-RL}} using two multi-agent RL environments and three different RL algorithms:
\begin{itemize}
    \item \textit{Multiagent Particle} (Predator Prey scenario) \cite{lowe2017multi}: {A} light environment with continuous observations and a discrete action space.
    {Predator-Prey scenario was evaluated }with three predators and a single prey. In this scenario, the prey is faster and aims to avoid being caught by predators.  
    The prey is positively rewarded when escaping predators, and negatively rewarded when caught, and {vice versa} for predators. Both types of agents are rewarded negatively when trying to overpass the screen boundaries or hitting an obstacle. As it was done by the authors of Multiagent Particle \cite{lowe2017multi}, the prey {was trained} using the DDPG \cite{lillicrap2015continuous} algorithm \footnote{DDPG implementation from repository \url{https://github.com/openai/maddpg/}} whereas predators were trained using MADDPG \cite{lowe2017multi}\footnote{MADDPG implementation: \url{https://github.com/openai/maddpg/}}. For this scenario, 5 different models  with the same hyperparameters {were trained} in order to obtain meaningful values despite the stochastic nature of MADDPG, as suggested in \cite{henderson2017deep} and detailed in the Appendix.

    \item \textit{Sequential Social Dilemmas} (\textit{Harvest} scenario) \cite{leibo2017multiagent}: {Another} light environment with continuous observations and discrete actions that proposes scenarios {emphasizing} social interactions and cooperation between agents. In the \textit{Harvest} scenario, agents must cooperate to harvest the maximum number of apples, while being careful to not ``kill" apple trees by collecting all apples they contain, as this would prevent this tree from spawning apples {further}. As it was done by the authors of Sequential Social Dilemmas open source implementation of the environment \cite{SSDOpenSource}, all agents were trained using the Asynchronous Advantage Actor-Critic (A3C) \cite{mnih2016asynchronous}\footnote{A3C implementation: \url{https://github.com/ray-project/ray/tree/master/rllib}} algorithm. The best reported hyperparameters and training protocol {showcased} in \cite{SSDOpenSource, DBLP:journals/corr/abs-1810-08647} {were used}. 
\end{itemize}

Two experiments {were conducted on the Predators-Prey scenario: 1) Using default settings provided by the authors of the environment} \cite{lowe2017multi} (to verify RQ1, RQ2 and RQ3), and 2) with different speeds for each predator (to further confirm RQ1 and RQ2). Speeds used on each experiment are presented in Table \ref{tab:experiments}.
For \textit{Harvest}, three experiments {were also conducted}: {first,} {the} Shapley {values} of a simple model trained with 6 agents using the default settings suggested by \cite{SSDOpenSource, DBLP:journals/corr/abs-1810-08647} {were computed} to further investigate RQ1 and, especially, {determine} the minimal number of agents required for this task. {Then,} {Shapley values were recomputed} using the same model but modified according to information extracted from the Shapley analysis obtained during the first experiment, in order to confirm the validity of such information. 
In addition, social outcome metrics from \cite{perolat2017multiagent} {are reported} to have a more fine-grained view of the payout notion (instead of merely a global reward). 
The following metrics {were implemented}:
\begin{itemize}
    \item \textit{Efficiency} (Eq. \ref{eq:efficiency}): {Measures} the total sum of all rewards obtained by all agents.
    \item \textit{Equality} (Eq. \ref{eq:equality}): {Measures} the statistical dispersion intended to represent inequality (Gini coefficient \cite{gini1912variabilita}).
    \item \textit{Sustainability} (Eq. \ref{eq:sustainability}): {Defined} as the average time $t_s \in t$ at which  rewards are collected. 
\end{itemize}

Considering $N$ independent agents, let $\{r_t^i | t = 1, . . . , T \}$ be the sequence of rewards obtained by
the i-th agent over an episode of duration $T$ timesteps. Its return is given by $R^i = \sum_{t=1}^{T}r^{i}_t$.
Thus, the equations describing the social metrics are as follows: 
\begin{equation}
\small
    \text{Efficiency U} = \mathbb{E}[\frac{\sum_{i=1}^{N} R^{i}}{T}] 
    \label{eq:efficiency}
\end{equation}
\begin{equation}
\small
    \text{Equality E} = 1 - \frac{\sum_{i=1}^{N} \sum_{j=1}^{N} |R^{i}-R^{j}|}{2N\sum_{i=1}^{N} R^{i}}
    \label{eq:equality}
\end{equation}
\begin{equation}
\small
    \text{Sustainability S} =\mathbb{E}[\frac{1}{N}\sum_{i=1}^{N}t^{i}], \text{where }  t^{i} = \mathbb{E}[t | r_t^{i} > 0] 
    \label{eq:sustainability}
\end{equation}

Finally, additional experiments {were} executed on different \textit{Harvest} checkpoints to explore how agents cooperate. Each experiment {will be detailed in the subsections below}. 
 
\vspace{0.5em}
\begin{table}[ht!]
 \caption{{Speed settings for each agent} used in Experiment 1 and 2 (RQ1, RQ2, RQ3), conducted on the Predator-Prey setting of Multiagent Particle \cite{lowe2017multi}.}
 \centering
\scalebox{0.7}{
\begin{tabular}{llllll}
  \textbf{} & \textbf{Prey} & \textbf{Pred. 1 (slow)} & \textbf{Pred. 2 (medium)} & \textbf{Pred. 3 (fast)}\\
  \midrule
  Speed in exp. \#1 & 1.3 & 1.0 & 1.0 & 1.0 \\
  Speed in exp. \#2 & 1.3 & 0.2 & 0.8 & 2.0 \\
\end{tabular}
}
    \label{tab:experiments}
\end{table}

\begin{table}[ht!]
    \small
    \caption{Table reporting Shapley {value computation times} for the Harvest \cite{SSDOpenSource} experiments (5 agents) run on a 6-core AMD Ryzen 5 5600X CPU, using Algorithm \ref{algo:1}. Computation time grows proportionally to $M$.}
    \centering
    \begin{tabular}{llll}
        \textit{M} & 100 & 500 & 1000 \\
         \textit{Computation time} & $\approx 1$ hour & $\approx 5$ hour & $\approx 10$ hour \\
    \end{tabular}
    \label{tab:computation_time}
\end{table}

\subsection{Experiment 1: Agents with Identical Settings in Multiagent Particle}
\label{subsec:exp_1}
\subsubsection{Environment Settings}
\label{subsec:exp1_settings}

The goal of this experiment is to {answer} RQ1 and RQ2 (Section \ref{sec:intro}), {i.e., whether the Shapley values of predator agents correlate with the number of times they catch the prey, and whether these are a close approximation to the exact Shapley values.}

First, when training a model of {three predators using default} settings on the Predator-Prey scenario, {one} could think that each predator should provide a similar contribution, as predators do not have significant differences on their speed, action space or training method. However, in contrast {to} this assumption, statistics in Figure \ref{fig:comp_perf_exp1} show that the performance of each predator agent ({i.e.,} the number of times each predator {catches} the prey) varies significantly: Predator 3 has a larger contribution than Predator 1, which {has} a higher contribution than Predator 0. In fact, the trained MADDPG model developed a strategy on which Predator 3 and Predator 2 perform better than Predator 0 at catching the prey. This can be explained by the fact that MADDPG provides the same reward to all agents, instead of only rewarding the {highest-contributing agent. This can} bias the training, as explained in \cite{wang2019shapley}. On the contrary, rewarding only the agent who contributed the most does not highlight or {recognize team strategies where the contribution of every agent was critical to the shared payout.}

\begin{figure}[ht!]
     \centering
     \includegraphics[width=0.4\textwidth]{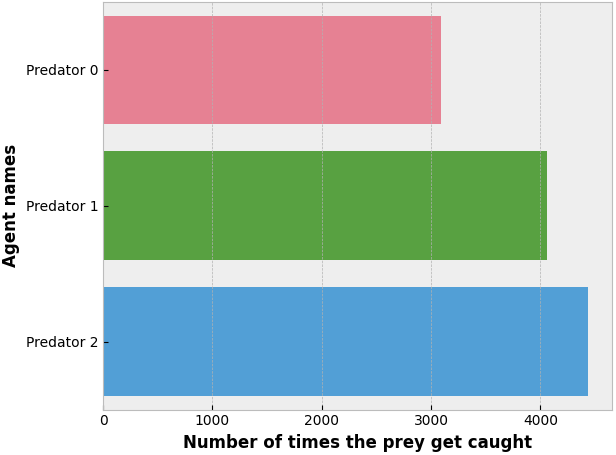}
     \includegraphics[width=0.4\textwidth]{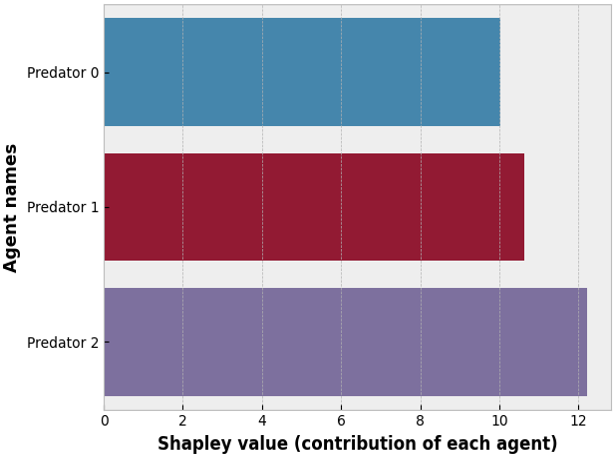}
    \caption{Predator-Prey environment: {Agents are endowed with the same speed.} Left plot shows predator agents mean performance comparison out of 10,000 sample episodes over 5 different models (run 2,000 sample episodes each) while right plot presents the Monte Carlo estimation of Shapley values obtained for each predator agent (M=1,000; ``random" player exclusion method; mean over 5 models).
    }
    \label{fig:comp_perf_exp1}
\end{figure}

\subsubsection{Shapley Values Analysis}
Figure \ref{fig:exp1_true_shap} shows Shapley values computed for each agent on each of the five models, over 1,000 sample episodes per model. This can be observed in a more convenient way in Figure \ref{fig:comp_perf_exp1}, which shows Shapley values computed for each agent on a single model. As hypothesized, the {decreasing} order of agents' Shapley values is the following: Predator 2, Predator 1 and Predator 0. Thus, this first experiment supports RQ1, since Shapley values are able to {correctly map contributions to agents usefulness} in a cooperative multi-agent setting.

\subsubsection{Comparison of Approximated Shapley Values with 
the {Exact} Shapley Values} 
\label{comp_real_exp1}
In this subsection, {the} Monte Carlo approximation of Shapley values {is compared with the exact} computation of Shapley values to verify RQ2. For that, Eq. \ref{eq:2} {is used} to compute Shapley values: marginal contributions are estimated as the mean global reward obtained by the coalition of players over a high number of episodes{--1,000 episodes in this experiment--using the player exclusion mechanisms in Section \ref{sect:mcshap}. Indeed, a large number of samples is needed for each coalition due to the stochastic nature of the environment, which leads to high variance in results. {As depicted in Fig. \ref{fig:exp1_true_shap}, Shapley values estimated by Monte Carlo} sampling are very close to the real Shapley values. {The difference of only 5\% on average for all agents and models therefore supports RQ2. In addition, MC approach} is simpler to implement and with a complexity that does not grow exponentially with the number of agents in the RL environment.}

\begin{figure}[htbp!]
    \centering
    \includegraphics[width=0.5\textwidth]{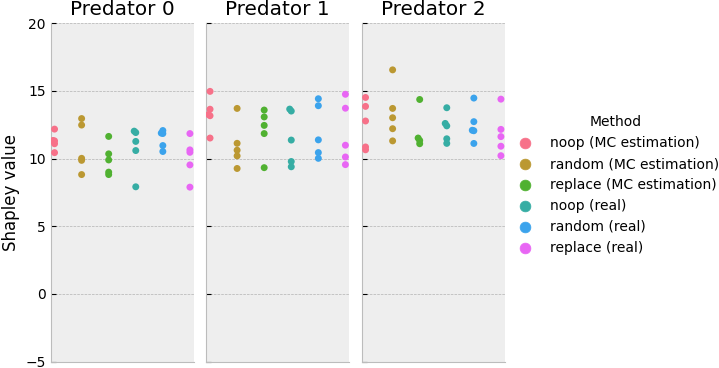}
    \caption{Predator-Prey environment, same agents' speeds. Comparison of MC approximation of Shapley values with the exact Shapley values.
    Each point relates to one of the 5 different models run.}
    \label{fig:exp1_true_shap}
\end{figure}

\subsection{Experiment 2: Introducing Variations in Agents Speeds in Multiagent Particle}
\label{subsec:exp_speeds}
\subsubsection{Environment Settings}
\label{subsec:exp2_settings}

{This experiment introduces} variations in the agent settings with the objective of disturbing the actual distribution of contributions between agents obtained in Experiment 1 and see if, as claimed in RQ1, this change will be reflected in the computed Shapley Values ({i.e.,} the aim is to ensure that Shapley values correlate with the observed contributions). Thus, the speed of each predator will differ from the default one {and so, it can be expected} that the faster agent will catch the prey more often and contribute the most to the global reward. In this manner, our goal is to obtain a clear hierarchy between agents that correlates with Shapley Values so they reflect the agents' observed behavior coherently. 
The following speeds {are arbitrarily set} for each agent: Predator 0 (slow): 0.2, Predator 1 (medium): 0.8, Predator 2 (fast): 2.0.
Statistics in Figure \ref{fig:comp_perf_exp2} show the performance of each predator agent in terms of the number of times each predator catches the prey.
As expected, the faster agent (Predator 3) {presents} a higher contribution than Predator 1, which {exhibits} a higher contribution than Predator 0 (the slowest). More precisely, the ranking in speed is reflected in the ranking of contributions {, which is the ideal setting to test if the assumption in RQ1 is valid.} In addition, the real Shapley values for these settings {were also computed} in order to further verify RQ2.

\begin{figure}[htbp!]
    \centering
    \includegraphics[width=0.4\textwidth]{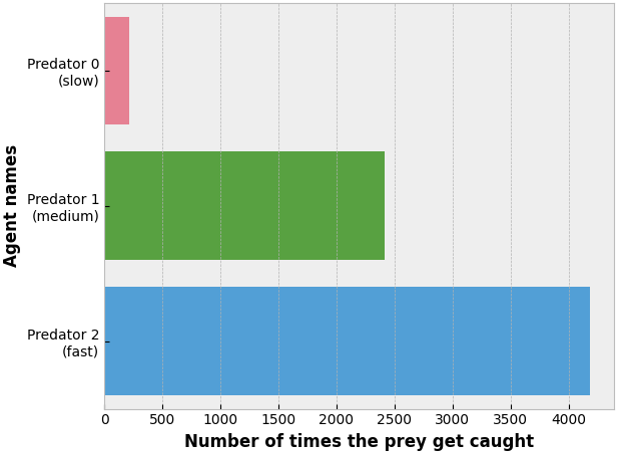}
    \includegraphics[width=0.4\textwidth]{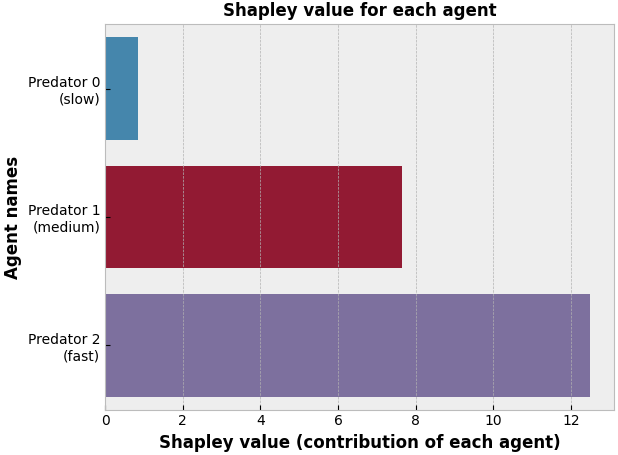}
    \caption{Predator-Prey environment, variable {agents'} speeds. The left plot shows predator {agents'} performance comparison for 10,000 sample episodes and 5 different models (2,000 sample episodes for each of the five trained models) while the right plot presents the Monte Carlo estimation of Shapley values (M=1,000) for each predator agent ({averaged} over 5 model runs) with the \textit{``random"} player exclusion option.
    }
    \label{fig:comp_perf_exp2}
\end{figure}

\subsubsection{Shapley Values Analysis} 

Figure \ref{fig:exp2_true_shap} shows the Shapley values computed for each agent on each of the five models, over 1,000 sample episodes per model. This can be observed in a more convenient way in Figure \ref{fig:comp_perf_exp2}, which shows the Shapley values computed for each agent on a single model. Following RQ1, the order of agents' Shapley values is the following: Predator 2{,} Predator 1{,} and finally Predator 0. These Shapley values accurately correlate with the number of times each of the agents caught the prey (see Figure \ref{fig:comp_perf_exp2}). In addition, as {noticeable} in Fig. \ref{fig:shap_exp2_speeds}, when making a single {agent's} speed vary, its Shapley value grows proportionally: the faster the agent is, the higher its Shapley value is. As expected, it makes sense, since it can {more easily} catch the prey, {hence} contributing more to the overall payout. Thus, this experiment further supports RQ1, since Shapley values accurately {correlate to the exact} (observed) distribution of contributions. 

\begin{figure}[htbp!]
    \centering
    \includegraphics[width=0.4\textwidth]{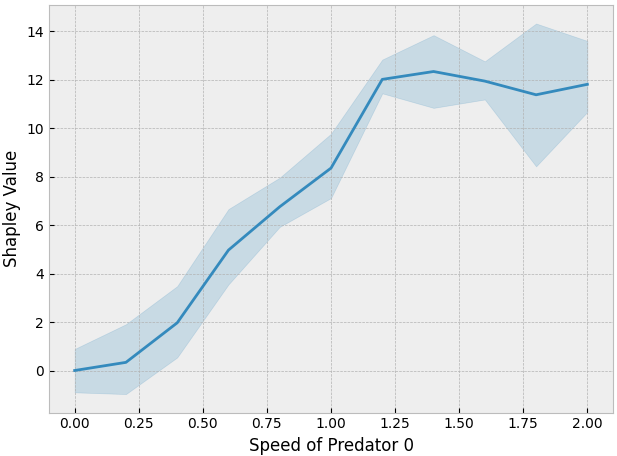}
    \caption{Predator-Prey environment, variable {agents'} speeds. Monte Carlo approximation of Shapley values (M=1,000) obtained by a single predator agent {with respect to} its speed. 
    }
    \label{fig:shap_exp2_speeds}
    
\end{figure}

\subsubsection{Comparison of Approximated Shapley Values with 
{Exact} Shapley Values} 
{In this section, the same experiment as described in Subsection \ref{comp_real_exp1} was conducted to further verify RQ2. The Monte Carlo approximation of Shapley values was compared with the 
{exact} or complete computation of Shapley values.} Here again, Fig. \ref{fig:exp2_true_shap} {showcases} that Shapley values estimated by Monte Carlo sampling (with $M=1,000$) are very close to the real ones with an average difference of 8\% between the approximated values and the real ones for Predator 1 and Predator 2, when computed with the same replacement method. However, this percentage reaches 53\% when considering only Predator 0, because its Shapley values are very close to 0 with a small standard deviation. {Therefore,} a small difference in value leads to a large difference {in} percentage. Therefore, approximated Shapley values are close to the real ones while being simpler to implement and with a complexity that does not grow with the number of agents in the RL environment, supporting {RQ2 again}.

\begin{figure}[htbp!]
    \centering
    \includegraphics[width=0.5\textwidth]{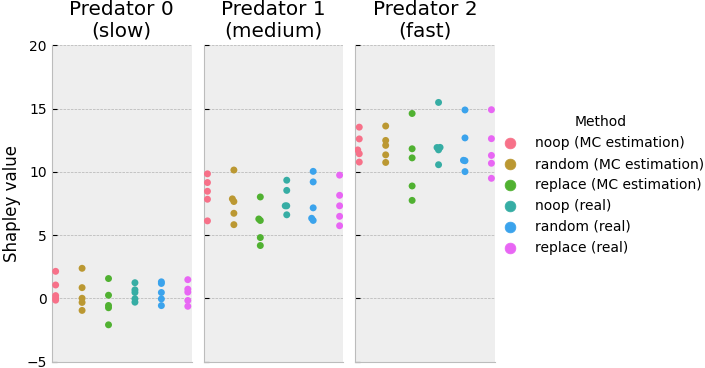}
    \caption{Predator-Prey environment, variable {agents'} speeds. Comparison of the Monte Carlo approximation of Shapley values with the real Shapley values. Each point {represents one of the five 
    model runs.}. 
    }
    \label{fig:exp2_true_shap}
\end{figure}

\subsection{Experiment 3: Explaining Agent {Contribution} in {the} \textit{Harvest} Environment}
\label{subsec:harvest}
\subsubsection{Environment Settings}
\label{subsec:exp4_settings}
In this experiment, the Monte Carlo Shapley Value computation method {is applied} to another multi-agent environment ({i.e.,} \textit{Harvest}). The goal here is to {test} 
this method in a more complex use case where {it could prove useful} to explain and extract insightful information from a trained model in which all agents share the same settings. This data {is leveraged in order} to attempt to further verify {RQ1} (presented in Section \ref{sec:intro}).

{Default} settings of \textit{Harvest} {were used}: 6 agents trying to collect apples on a pre-configured map. At first glance, this seems quite a large number of agents for a map this small (39*15, with 159 apples initially). Thus, {one} can hypothesize {that} some agents are superfluous, not contributing much to the global reward, and may even prevent other agents from elaborating effective strategies together, obstructing them in their movements. 

\subsubsection{Shapley Values Analysis}
\label{subsubsec:shapley_val_exp3}
First, Figure \ref{fig:shap_one_exp3} clearly {highlights the fact} that Agent 5 does not seem to bring much added value to the team ({i.e., its} Shapley value {is close} to 0) while {all other agents seem to contribute a near} equal amount to the global reward. In fact, while watching the agents {play}, {it appears} that Agent 5 is left unaccounted for, wandering on the map, not harvesting any apple and sometimes randomly hitting the map border, which grants {it} a negative reward. This may indicate that the default setting with 6 agents is 
{unnecessary or unproductive}: training only 5 agents could be enough to provide the same level of performance, {will require less hardware and be less time consuming}. Moreover, the fact that the first five agents contribute equally may {prove} that the \textit{A3C} \cite{mnih2016asynchronous} {model} found a satisfying solution to distribute tasks among agents, with the exception of Agent 5 who is left unaccounted for, maybe, {and reasonably,} because none of its actions could help increase the global reward. {In conclusion},
this synergy between agents, observed both in the obtained global reward and the {nearly} equal partition of Shapley values among agents, shows that agents are actually cooperating {with each other}.  

\begin{figure}[htbp!]
    \centering
    \includegraphics[width=0.4\textwidth]{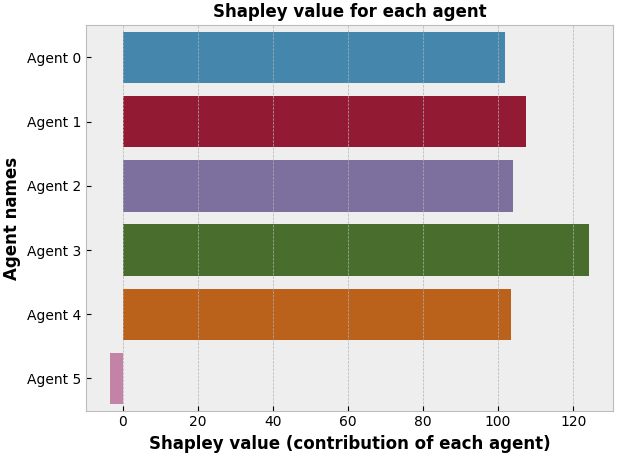}
    \includegraphics[width=0.4\textwidth]{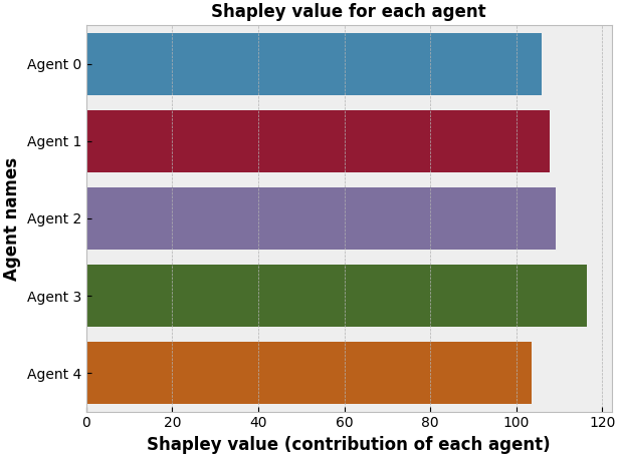}
    \caption{Harvest environment: The left plot shows the Monte Carlo estimation of Shapley values obtained for each agent (M=1,000, ``noop" action selection method).
    The right plot displays Shapley values computed {over agents with} the same settings but without Agent 5 {present}.}
    \label{fig:shap_one_exp3}
\end{figure}

\begin{figure}[htbp!]
    \centering
    \includegraphics[width=0.5\textwidth]{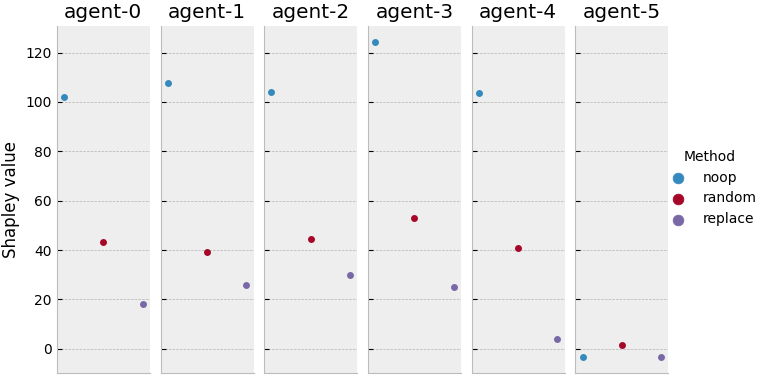}
    \caption{Harvest environment. Monte Carlo estimation of Shapley values (M=1,000) for each agent using each of the three agent substitution methods. The same settings were used for all agents.}
    \label{fig:harvest_shap}
\end{figure}

\subsubsection{Following The Insight Given by Shapley Values}

{In Subsection \ref{subsubsec:shapley_val_exp3}, the Shapley values analysis} indicated that Agent 5 does not contribute at all to solve the \textit{Harvest} environment. {Thus, it was} decided to re-run the Shapley value computation on the same model; this time with Agent 5 removed ({i.e.,} completely deactivated and not appearing in the environment map) to {check, whether, as suspected, the global reward and agents' contribution remained} the same. {Figure} \ref{fig:shap_one_exp3} {shows} that Shapley values distribution when Agent 5 is deactivated stays nearly identical to the previous setting {where Agent 5 is included in the game}. Some very minor variation in Shapley values can be attributed to stochasticity, {since each estimation of Shapley values does not yield the exact same results every time}. In addition, the global reward {also} remains stable at around 450 {reward units}.
This means that removing Agent 5 did not have any negative effect on the game and corroborates our 
{hypothesis} that its participation was not productive. {In conclusion, a valuable insight was successfully derived} from the analysis of Shapley values, strengthening the validity of {the hypothesis in} RQ1.

\subsubsection{Social Outcome Metrics: Analysis of the Social Behavior Between Agents}
\label{subsec:harvest_social}

{This section presents an analysis of how each social outcome metric (introduced in \cite{perolat2017multiagent}\footnote{The Peace metric, also presented in \cite{perolat2017multiagent}, {was not included} because it relies on a specific mechanism ({i.e., a} \textit{time-out} period during which a tagged agent cannot harvest apples anymore) that {was deemed as of limited utility by the authors, since} agents would learn very quickly not to use it.} (whose definitions are given in Subsection \ref{subsec:exp_context}) can be explained with our approach. The goal here is to further explore RQ1 and see {whether} agents leverage social learning and cooperate with each other {in practice}.} Note that the social 
and sustainability metrics (as originally defined in \cite{perolat2017multiagent}) require credit assignment to be computed in a per-agent basis, {while this} is not the case for Shapley values, which only {needs to be given} a global reward.

\begin{figure*}[htbp!]
    \centering
    \includegraphics[scale=0.36]{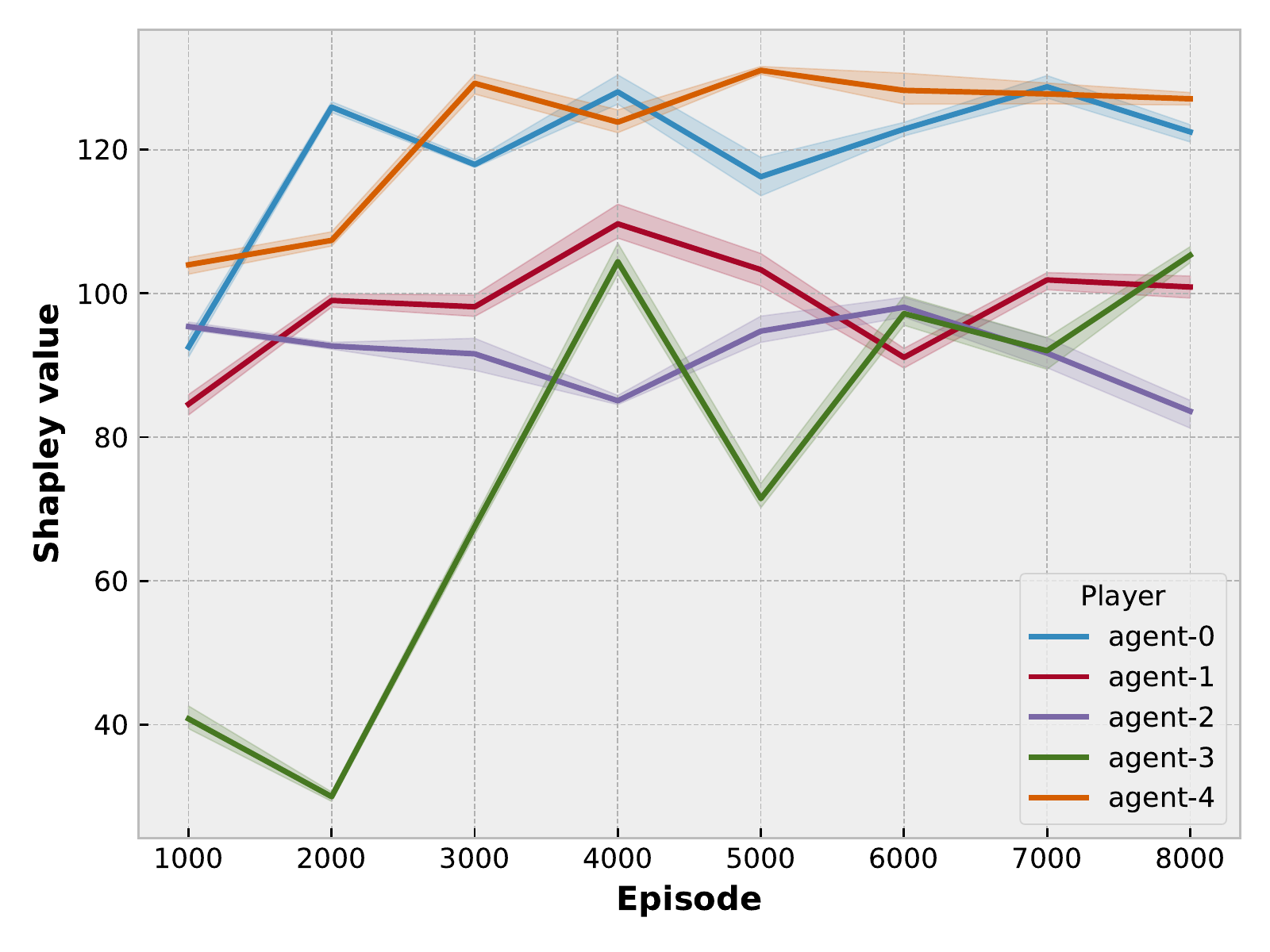}
    \includegraphics[scale=0.36]{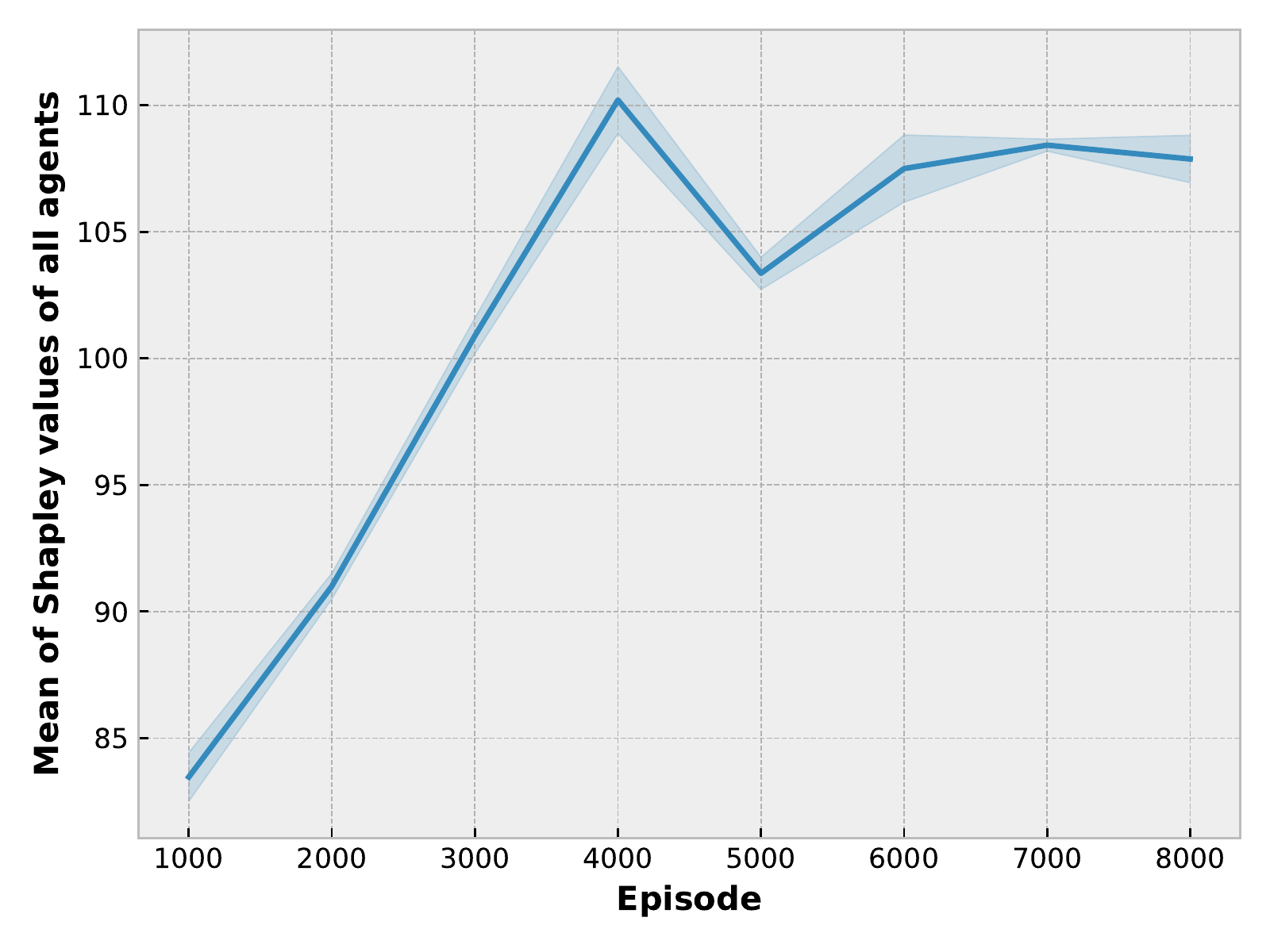}
    \includegraphics[scale=0.36]{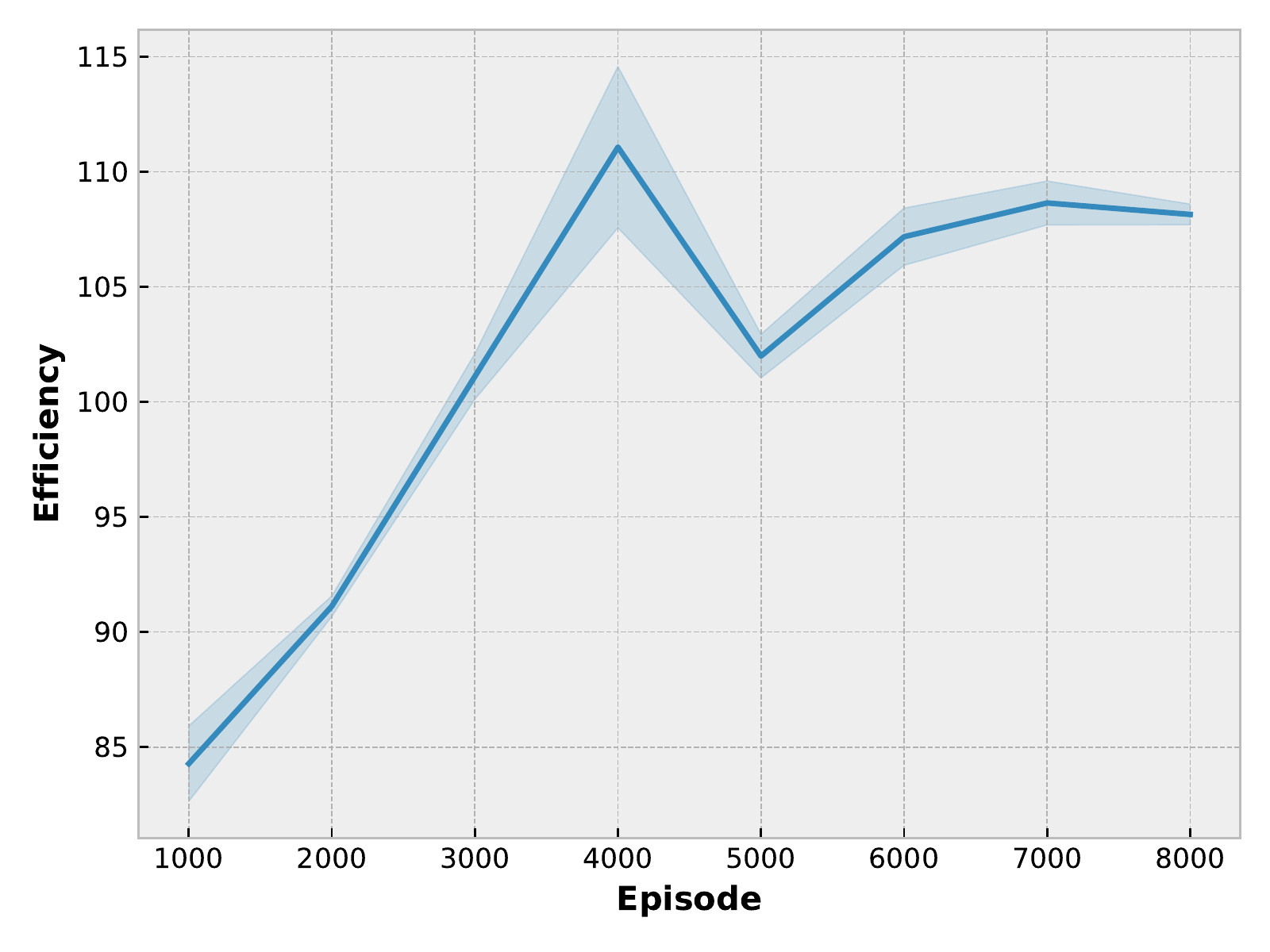}
    \includegraphics[scale=0.36]{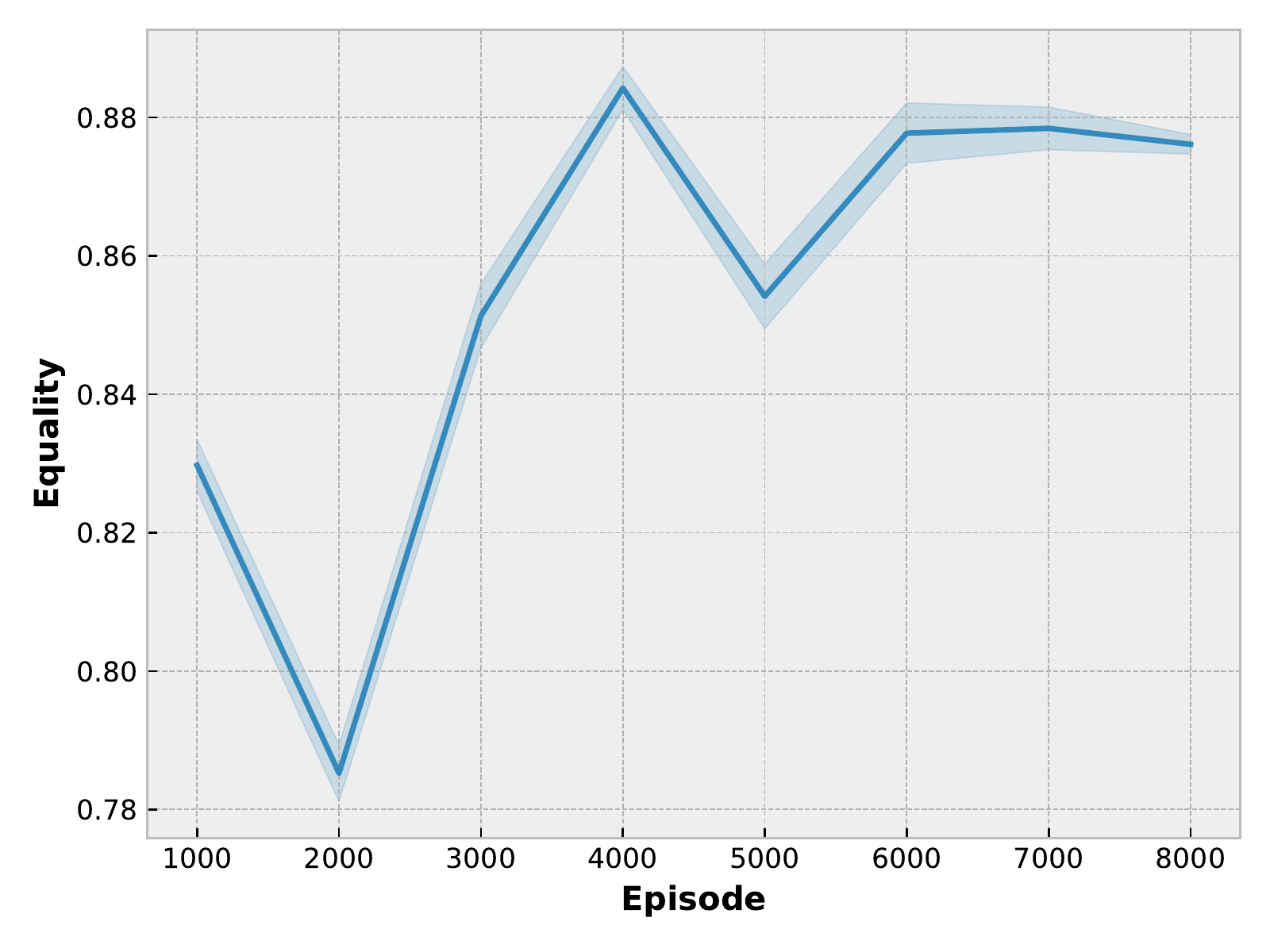}
    \includegraphics[scale=0.36]{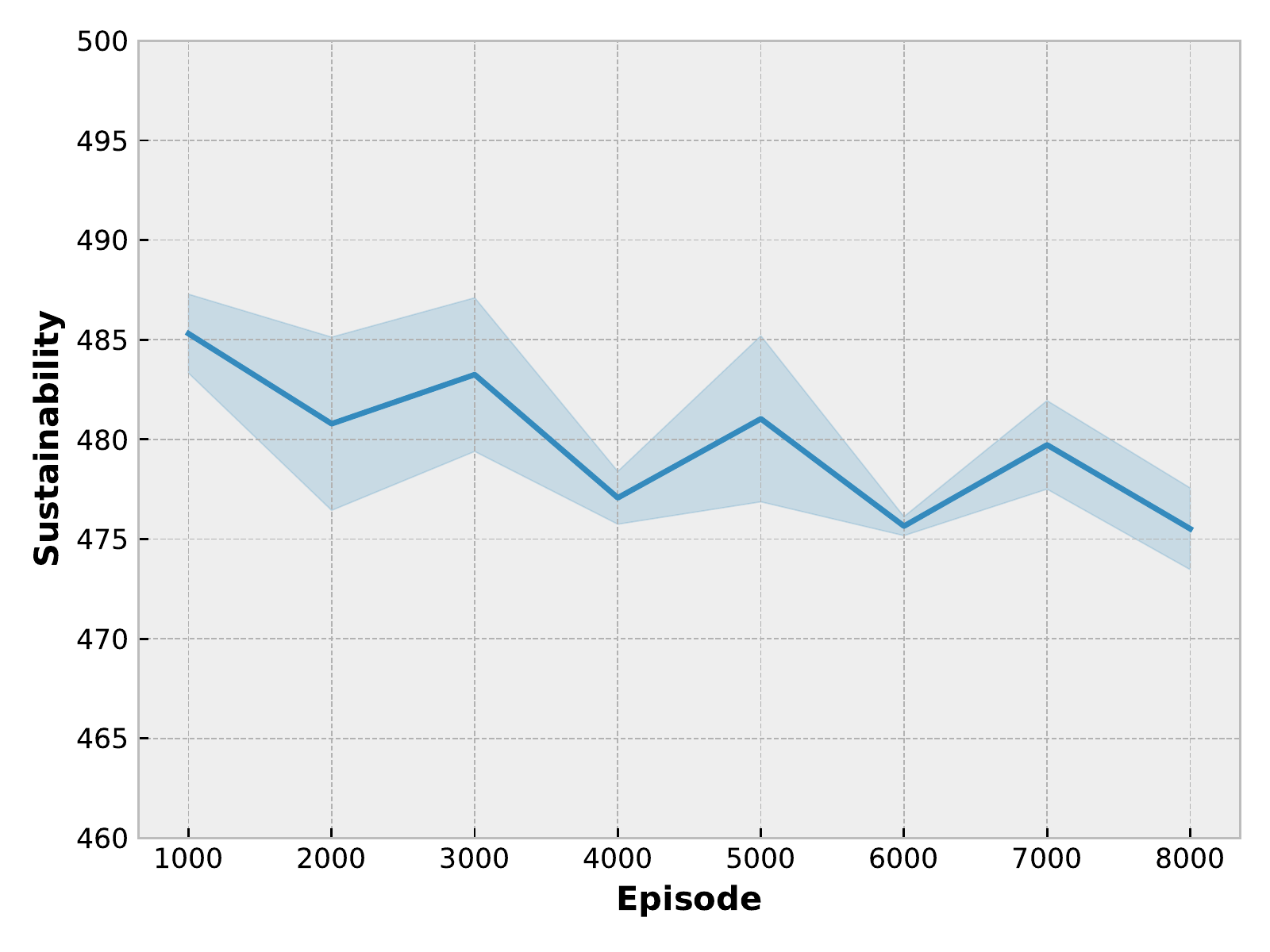}
    \caption{Harvest environment: Evolution of the Shapley values and social metrics from \cite{perolat2017multiagent}
    over different training episodes with A3C model. From top to bottom are displayed the Shapley values (\textit{noop} action selection method; {MC} estimation with M=500), the mean of those Shapley values of all agents, the efficiency, {equality, and sustainability metrics.}}
    \label{fig:social_metrics_harvest}
\end{figure*}
For {these} experiments, {over 100 training episodes were run, the three social outcome metrics presented above were computed, as well as} the Shapley values (using Monte Carlo approximation and M=500), at different steps of the training ({i.e.,} every 1,000 episodes), in order to analyze the evolution of those values during training. Results are presented in Figure \ref{fig:social_metrics_harvest}. As the Efficiency metric is nothing else than the mean of the Shapley values (in expectation), {it can be observed} with no surprise that the mean of the Shapley values of all agents has the same evolution than the {e}fficiency, and directly correlates with it. The slight differences, such as the peak at episode 4,000, are due to the stochastic nature of the environment, the choice of $M$ in Monte Carlo approximation and {number} of episodes used to compute the metrics.

{From previous experiment it can be concluded} that the mean of Shapley values of all agents is a metric that explains the agent's {a}fficiency and does not assume any credit assignment among agents. {As} stated in Equation \ref{eq:4}, a shared global reward is enough. 
Equality evolves in the same manner (with a peak at episode 4,000 and a drop at episode 5,000). This metric decreases a bit at episode 2,000: this fall is not captured by {the efficiency nor} the mean of Shapley Values. However, while looking at the Shapley values of each agent in Figure \ref{fig:social_metrics_harvest}, {it is clear} that {agent 3} obtains a lower reward than other agents. Using the Shapley values of all agents, {an explanation can thus be} that the decrease in equality at episode 2,000 is caused by agent 3, which contributes little to the global reward ({in comparison with other agents}). {It could not have} been {possible} to explain this behavior using {the equality metric} in a shared global reward context and herein the value of applying Shapley analysis in this context. {As Shapley values are computed using the global reward of the total team of agents, 
no link can be established} between {the sustainability metric and Shapley values. Especially because \textit{sustainability} is defined as the average time (\textit{time-step} in our use cases) at which rewards are collected, and this is independent from the value of these rewards.}

{This experiment showed} that Shapley values can effectively capture both {metrics; especially 1) if agents get high rewards (Efficiency) and 2) if rewards} are shared equally among all agents (Equality). However, Shapley values cannot tell if {agents are obtaining their} reward continuously ({i.e., displaying a} \textit{sustainable} {behaviour). 
Contrarily to the} social outcome metrics, Shapley values can be computed even though the reward is globally shared {among agents}. This is a huge advantage when the environment does not allow {precise} credit assignment.

\subsection{{Choosing a Player Exclusion Method}}
\label{subsec:player_exclusion}

In this subsection, {results gathered using the different player exclusion options} (defined in Section \ref{sect:mcshap}) during the three experiments above {are analyzed} in order to answer RQ3. 

When looking at Figures \ref{fig:exp1_true_shap}, \ref{fig:exp2_true_shap} and \ref{fig:harvest_shap}, {it becomes clear that the three player exclusion mechanisms lead to Shapley values that, when considered individually, are coherent for the excluded agent with respect to the others}.
However, in the referential of a single agent, the standard deviation between their respective values is important. In particular, there is a significant gap between \textit{noop} and the other two methods. {Figure \ref{fig:harvest_shap} shows there is} an average gap of 73.1 reward units between \textit{noop} and \textit{random\_player\_action}, while there is only an average gap of 20.5 reward units between \textit{random} and \textit{random\_player\_action}). This can be explained by the fact that randomly moving agents disturb the game {significantly} more than immobilized ones, as they can get negative rewards by hitting the map borders in \textit{Multiagent Particle} or \textit{killing} trees in \textit{Harvest} ({i.e., harvesting all} apples that are contiguous). Thus, when using \textit{random} or \textit{replace} {strategy}, a majority of coalitions are ``parasited" by these negative rewards that contribute {towards lowering} the global reward and lead to {an overall lower Shapley value} than the \textit{noop} method (as {observed} in Figures \ref{fig:exp1_true_shap}, \ref{fig:exp2_true_shap} and \ref{fig:harvest_shap}). {Therefore}, in that context, \textit{noop} action selection seems to be the most 
\textbf{faithful} method to get Shapley values assessing the agents' true contributions {closely, and free from random and unwanted} negative rewards. 
\vspace{-0.5em}

\section{Discussion}
\label{sec:discussion}
{This article} demonstrated the usefulness of Shapley values and their Monte Carlo approximation for explaining RL models in cooperative settings. 
{These values provide a form of explanation, } {i.e., continuous values that are understandable by researchers and developers, since they represent a portion} of the reward value of the agents team, partitioned according to each {agent's} contribution. They could also provide explanations {for the general public that may perceive them as an} intrinsic ``value" of each agent, 
{making them accountable for} the effectiveness of the system. Moreover, Shapley values could be a good way to detect biases in the training of an RL model, since they require {analyzing} the individual behavior of each agent and {this could} highlight disparities between their different strategies {and abilities}. 

Concerning the player exclusion method to replace missing agents from a coalition, \textit{noop} (no-operation) action seems to be the most neutral, {and} interaction-free method when the environment {offers this possibility, since methods using a substitution mechanism mandated by random-selection of actions} are prone to {get} high negative rewards and interfere in the game. {Social interaction between agents {was} also explored} and {this investigation showed} that Shapley values are able to effectively capture both \textit{efficiency} and \textit{equality} metrics, while {they are still} able to be computed even though the reward is globally shared between agents. This is a huge advantage when the environment does not enable fair individual-level credit assignment.
{In} consequence, {it can be asserted} that Shapley Values are an effective way to explain the contributions of RL agents, and, to some extent, the relationships between them.

However, our approach is limited to multi-agent cooperative RL and, in its current form, cannot be applied to competitive and single-agent models. In addition, it cannot be used to explain an agent's actions, {their sustainability in time}, nor explain a specific episode of interest, as it only provides an average metric for the contribution of each agent in a cooperative game, with the total of Shapley values corresponding to the mean global reward of the grand coalition ({i.e.,} the one containing all agents). Thus, it must be considered as a way to get a first {ranking of contributions of agents in} a model.
Finally, while {the Monte Carlo method to estimate Shapley values} (see Section \ref{sect:mcshap}) is more 
{efficient} than computing the exact Shapley values, it still remains time consuming. {Future work should seek to keep {accurate value estimation of SHAP values} while accelerating their computational approximation. }

\section{Conclusion and Future Work}
\label{sec:conclusion}
{The three research questions} 
{were positively answered}, with experiments conducted in two socially challenging multi-agent RL environments (\textit{Harvest} from Sequential Social Dilemmas \cite{SSDOpenSource, leibo2017multiagent} and \textit{Particle Multiagent} \cite{lowe2017multi}) and two different RL algorithms (MADDPG \cite{lowe2017multi} and A3C \cite{mnih2016asynchronous}).
{Experiments} showed that{the computation of} Shapley values could be a potential breakthrough elucidating understanding towards attaining multi-agent XRL environments. They can efficiently assess the contribution of agents to the global reward in cooperative settings. {They also provide} insightful information about the agents' {behaviour} and their social interactions.

Nonetheless, numerous issues remain to {be explained} in future work. Different interpretations of Shapley values to {further explain deep RL issues must be explored} to increase the levels of explanation granularity.
Robustness {and reproducibility remain} a critical issue for XRL (and XAI in a more general sense), and other statistical methods could prove very useful for that purpose, as presented in \cite{huber2004robust, maronna2019robust, hampel2011robust} ({e.g.,} Winsorised or trimmed estimators). Moreover, Shapley values could also be combined with a robust model selection measure (such as the Lorenz Zonoids 
\cite{GiudiciShapleyLorentz}). Besides, Shapley values or other additive and {non-additive} methods could be used {not only} to explain the roles taken by agents when learning a policy to achieve a collaborative task, but also to detect defects in agents while training, or {in} the fed data. 
Furthermore, the dynamic nature of RL (vs. the static settings of most ML models where only a single data point needs to be explained) {could be taken into account in order to create a novel approach that evaluates the contributions of agents through time (e.g. during evaluation time}). Here, ``temporal" Shapley values could be approximated with a model as in \cite{wang2019shapley}. However, one of the main advantages of SHAP being a post-hoc XAI method ({i.e.,} being agnostic to the RL algorithm) {would be lost}, as the Shapley prediction model would 
be {dependent on 
the 
policy learning} model used. Finally, a different contribution ranking scheme than the one presented in Section \ref{sect:mcshap} {could also be proposed} ({i.e., accounting for more complex objective metrics to better highlight the }order of importance in the team). For instance, each set of observations per agent {could be ranked in order of quality or average \textit{didactic} importance 
in order to assess the learning agents more fairly, with respect to the quality of the data they were exposed to.}

\ifCLASSOPTIONcaptionsoff
  \newpage
\fi



\bibliographystyle{IEEEtran}
\bibliography{IEEEabrv,refs.bib}

\begin{thebibliography}{10}
\providecommand{\url}[1]{#1}
\csname url@samestyle\endcsname
\providecommand{\newblock}{\relax}
\providecommand{\bibinfo}[2]{#2}
\providecommand{\BIBentrySTDinterwordspacing}{\spaceskip=0pt\relax}
\providecommand{\BIBentryALTinterwordstretchfactor}{4}
\providecommand{\BIBentryALTinterwordspacing}{\spaceskip=\fontdimen2\font plus
\BIBentryALTinterwordstretchfactor\fontdimen3\font minus
  \fontdimen4\font\relax}
\providecommand{\BIBforeignlanguage}[2]{{%
\expandafter\ifx\csname l@#1\endcsname\relax
\typeout{** WARNING: IEEEtran.bst: No hyphenation pattern has been}%
\typeout{** loaded for the language `#1'. Using the pattern for}%
\typeout{** the default language instead.}%
\else
\language=\csname l@#1\endcsname
\fi
#2}}
\providecommand{\BIBdecl}{\relax}
\BIBdecl

\bibitem{mnih2016asynchronous}
V.~Mnih, A.~P. Badia, M.~Mirza, A.~Graves, T.~Lillicrap, T.~Harley, D.~Silver,
  and K.~Kavukcuoglu, ``{Asynchronous Methods for Deep Reinforcement
  Learning},'' in \emph{International Conference on Machine Learning}.\hskip
  1em plus 0.5em minus 0.4em\relax PMLR, 2016, pp. 1928--1937.

\bibitem{espeholt2018impala}
L.~Espeholt, H.~Soyer, R.~Munos, K.~Simonyan, V.~Mnih, T.~Ward, Y.~Doron,
  V.~Firoiu, T.~Harley, I.~Dunning \emph{et~al.}, ``{Impala: Scalable
  Distributed Deep-RL with Importance Weighted Actor-Learner Architectures},''
  in \emph{International Conference on Machine Learning}.\hskip 1em plus 0.5em
  minus 0.4em\relax PMLR, 2018, pp. 1407--1416.

\bibitem{ribeiro2016i}
M.~T. Ribeiro, S.~Singh, and C.~Guestrin, ``{`Why Should I Trust You?'
  Explaining the Predictions of any Classifier},'' in \emph{Proceedings of the
  22nd ACM SIGKDD International Conference on Knowledge Discovery and Data
  Mining}, 2016, pp. 1135--1144.

\bibitem{el2021multilayer}
S.~El-Sappagh, J.~M. Alonso, S.~R. Islam, A.~M. Sultan, and K.~S. Kwak, ``{A
  Multilayer Multimodal Detection and Prediction Model based on Explainable
  Artificial Intelligence for Alzheimer’s Disease},'' \emph{Scientific
  Reports}, vol.~11, no.~1, pp. 1--26, 2021.

\bibitem{arnout2019towards}
U.~Schlegel, H.~Arnout, M.~El-Assady, D.~Oelke, and D.~A. Keim, ``{Towards A
  Rigorous Evaluation of XAI Methods On Time Series},'' in \emph{IEEE/CVF
  International Conference on Computer Vision Workshop (ICCVW)}, 2019, pp.
  4197--4201.

\bibitem{HEUILLET2021106685}
\BIBentryALTinterwordspacing
A.~Heuillet, F.~Couthouis, and N.~Díaz-Rodríguez, ``{Explainability in Deep
  Reinforcement Learning},'' \emph{Knowledge-Based Systems}, vol. 214, p.
  106685, 2021. [Online]. Available:
  \url{http://www.sciencedirect.com/science/article/pii/S0950705120308145}
\BIBentrySTDinterwordspacing

\bibitem{puiutta2020explainable}
E.~Puiutta and E.~M. Veith, ``{Explainable Reinforcement Learning: A Survey},''
  in \emph{International Cross-Domain Conference for Machine Learning and
  Knowledge Extraction}.\hskip 1em plus 0.5em minus 0.4em\relax Springer, 2020,
  pp. 77--95.

\bibitem{madumal2019explainable}
P.~Madumal, T.~Miller, L.~Sonenberg, and F.~Vetere, ``{Explainable
  Reinforcement Learning through a Causal Lens},'' in \emph{Proceedings of the
  AAAI Conference on Artificial Intelligence}, vol.~34, no.~03, 2020, pp.
  2493--2500.

\bibitem{atarivisualizing}
S.~Greydanus, A.~Koul, J.~Dodge, and A.~Fern, ``{Visualizing and Understanding
  Atari Agents},'' in \emph{International Conference on Machine
  Learning}.\hskip 1em plus 0.5em minus 0.4em\relax PMLR, 2018, pp. 1792--1801.

\bibitem{lundberg2017unified}
S.~M. Lundberg and S.-I. Lee, ``{A Unified Approach to Interpreting Model
  Predictions},'' in \emph{Proceedings of the 31st International Conference on
  Neural Information Processing Systems}, 2017, pp. 4768--4777.

\bibitem{shapley_1953}
L.~Shapley, ``{A Value for N-Person Games},'' \emph{Contributions to the Theory
  of Games}, no.~28, pp. 307--317, 1953.

\bibitem{alain21collaborative}
A.~Andres, E.~Villar-Rodriguez, A.~D. Martinez, and J.~Del~Ser, ``Collaborative
  exploration and reinforcement learning between heterogeneously skilled agents
  in environments with sparse rewards,'' in \emph{2021 International Joint
  Conference on Neural Networks (IJCNN)}, 2021, pp. 1--10.

\bibitem{hardin2009tragedy}
G.~Hardin, ``{The Tragedy of the Commons},'' \emph{Journal of Natural Resources
  Policy Research}, vol.~1, no.~3, pp. 243--253, 2009.

\bibitem{chica2021collective}
M.~Chica, J.~M. Hern{\'a}ndez, and J.~Bulchand-Gidumal, ``{A Collective Risk
  Dilemma for Tourism Restrictions under the COVID-19 Context},''
  \emph{Scientific Reports}, vol.~11, no.~1, pp. 1--12, 2021.

\bibitem{lowe2017multi}
\BIBentryALTinterwordspacing
R.~Lowe, Y.~Wu, A.~Tamar, J.~Harb, P.~Abbeel, and I.~Mordatch, ``{Multi-Agent
  Actor-Critic for Mixed Cooperative-Competitive Environments},'' \emph{Neural
  Information Processing Systems (NIPS)}, 2017. [Online]. Available:
  \url{https://arxiv.org/pdf/1706.02275.pdf}
\BIBentrySTDinterwordspacing

\bibitem{leibo2017multiagent}
J.~Z. Leibo, V.~Zambaldi, M.~Lanctot, J.~Marecki, and T.~Graepel,
  ``{Multi-Agent Reinforcement Learning in Sequential Social Dilemmas},'' in
  \emph{Proceedings of the 16th Conference on Autonomous Agents and MultiAgent
  Systems}, ser. AAMAS '17.\hskip 1em plus 0.5em minus 0.4em\relax
  International Foundation for Autonomous Agents and Multiagent Systems, 2017,
  p. 464–473.

\bibitem{Barredo20}
A.~B. Arrieta, N.~D{\'\i}az-Rodr{\'\i}guez, J.~Del~Ser, A.~Bennetot, S.~Tabik,
  A.~Barbado, S.~Garc{\'\i}a, S.~Gil-L{\'o}pez, D.~Molina, R.~Benjamins
  \emph{et~al.}, ``{Explainable Artificial Intelligence (XAI): Concepts,
  Taxonomies, Opportunities and Challenges Toward Responsible AI},''
  \emph{Information Fusion}, vol.~58, pp. 82--115, 2020.

\bibitem{ndousse2021emergent}
K.~K. Ndousse, D.~Eck, S.~Levine, and N.~Jaques, ``Emergent social learning via
  multi-agent reinforcement learning,'' in \emph{International Conference on
  Machine Learning}.\hskip 1em plus 0.5em minus 0.4em\relax PMLR, 2021, pp.
  7991--8004.

\bibitem{perolat2017multiagent}
J.~Perolat, J.~Z. Leibo, V.~Zambaldi, C.~Beattie, K.~Tuyls, and T.~Graepel,
  ``{A Multi-Agent Reinforcement Learning Model of Common-Pool Resource
  Appropriation},'' in \emph{Proceedings of the 31st International Conference
  on Neural Information Processing Systems}, 2017, pp. 3646--3655.

\bibitem{hughes2018inequity}
E.~Hughes, J.~Leibo, M.~Phillips, K.~Tuyls, E.~Duenez-Guzman, A.~Castaneda,
  I.~Dunning, T.~Zhu, K.~McKee, R.~Koster \emph{et~al.}, ``{Inequity Aversion
  Improves Cooperation in Intertemporal Social Dilemmas},'' in \emph{Advances
  in Neural Information Processing Systems 31}, vol.~31.\hskip 1em plus 0.5em
  minus 0.4em\relax Neural Information Processing Systems Foundation, Inc.,
  2018, pp. 1--11.

\bibitem{Staniak18}
\BIBentryALTinterwordspacing
M.~Staniak and P.~Biecek, ``{Explanations of Model Predictions with live and
  breakDown Packages},'' \emph{{The R Journal}}, vol.~10, no.~2, pp. 395--409,
  2018. [Online]. Available: \url{10.32614/RJ-2018-072}
\BIBentrySTDinterwordspacing

\bibitem{GiudiciShapleyLorentz}
P.~Giudici and E.~Raffinetti, ``{Shapley-Lorenz Decompositions in eXplainable
  Artificial Intelligence},'' \emph{SSRN Electronic Journal}, 01 2020.

\bibitem{mai2020adversarial}
D.~Shim, Z.~Mai, J.~Jeong, S.~Sanner, H.~Kim, and J.~Jang, ``{Online
  Class-Incremental Continual Learning with Adversarial Shapley Value},'' in
  \emph{Proceedings of the AAAI Conference on Artificial Intelligence},
  vol.~35, no.~11, 2021, pp. 9630--9638.

\bibitem{Sundararajan20}
M.~Sundararajan and A.~Najmi, ``{The Many Shapley Values for Model
  Explanation},'' in \emph{International Conference on Machine Learning}.\hskip
  1em plus 0.5em minus 0.4em\relax PMLR, 2020, pp. 9269--9278.

\bibitem{wang2019shapley}
J.~Wang, Y.~Zhang, T.-K. Kim, and Y.~Gu, ``{Shapley Q-value: A Local Reward
  Approach to Solve Global Reward Games},'' in \emph{Proceedings of the AAAI
  Conference on Artificial Intelligence}, vol.~34, no.~05, 2020, pp.
  7285--7292.

\bibitem{kiran2021deep}
B.~R. Kiran, I.~Sobh, V.~Talpaert, P.~Mannion, A.~A. Al~Sallab, S.~Yogamani,
  and P.~P{\'e}rez, ``{Deep Reinforcement Learning for Autonomous Driving: A
  Survey},'' \emph{IEEE Transactions on Intelligent Transportation Systems},
  2021.

\bibitem{Nguyen_rl_robotics}
H.~Nguyen and H.~La, ``{Review of Deep Reinforcement Learning for Robot
  Manipulation},'' in \emph{Third IEEE International Conference on Robotic
  Computing (IRC)}, 2019, pp. 590--595.

\bibitem{lesort2020continual}
T.~Lesort, V.~Lomonaco, A.~Stoian, D.~Maltoni, D.~Filliat, and
  N.~D{\'\i}az-Rodr{\'\i}guez, ``{Continual Learning for Robotics: Definition,
  Framework, Learning Strategies, Opportunities and Challenges},''
  \emph{Information Fusion}, vol.~58, pp. 52--68, 2020.

\bibitem{lee2021joint}
\BIBentryALTinterwordspacing
D.~Lee, N.~Jaques, J.~C. Kew, D.~Eck, D.~Schuurmans, and A.~Faust, ``Joint
  attention for multi-agent coordination and social learning,'' \emph{CoRR},
  vol. abs/2104.07750, 2021. [Online]. Available:
  \url{https://arxiv.org/abs/2104.07750}
\BIBentrySTDinterwordspacing

\bibitem{ndousse2020multi}
K.~Ndousse, D.~Eck, S.~Levine, and N.~Jaques, ``Multi-agent social
  reinforcement learning improves generalization,'' \emph{arXiv e-prints}, pp.
  arXiv--2010, 2020.

\bibitem{DUFFY2009785}
\BIBentryALTinterwordspacing
J.~Duffy and J.~Ochs, ``{Cooperative Behavior and the Frequency of Social
  Interaction},'' \emph{Games and Economic Behavior}, vol.~66, no.~2, pp. 785
  -- 812, 2009, special Section In Honor of David Gale. [Online]. Available:
  \url{http://www.sciencedirect.com/science/article/pii/S0899825608001395}
\BIBentrySTDinterwordspacing

\bibitem{Colman2003}
A.~M. Colman, ``{Cooperation, Psychological Game Theory, and Limitations of
  Rationality in Social Interaction},'' \emph{The Behavioral and Brain
  Sciences}, vol. 26(2), p. 139–198, 2003.

\bibitem{DBLP:journals/corr/abs-1810-08647}
N.~Jaques, A.~Lazaridou, E.~Hughes, C.~Gulcehre, P.~Ortega, D.~Strouse, J.~Z.
  Leibo, and N.~De~Freitas, ``{Social Influence as Intrinsic Motivation for
  multi-agent deep reinforcement learning},'' in \emph{International Conference
  on Machine Learning}.\hskip 1em plus 0.5em minus 0.4em\relax PMLR, 2019, pp.
  3040--3049.

\bibitem{ndousse2020sociallearning}
\BIBentryALTinterwordspacing
K.~Ndousse, D.~Eck, S.~Levine, and N.~Jaques, ``{Learning Social Learning},''
  in \emph{NeurIPS Workshop on Cooperative AI}, 2020. [Online]. Available:
  \url{https://arxiv.org/abs/2010.00581}
\BIBentrySTDinterwordspacing

\bibitem{Molnar19}
C.~Molnar, \emph{Interpretable Machine Learning}, 2019,
  \url{https://christophm.github.io/interpretable-ml-book/}.

\bibitem{roth_1988}
\BIBentryALTinterwordspacing
\emph{The Shapley Value: Essays in Honor of Lloyd S. Shapley}.\hskip 1em plus
  0.5em minus 0.4em\relax Cambridge University Press, 1988. [Online].
  Available: \url{http://www.library.fa.ru/files/Roth2.pdf}
\BIBentrySTDinterwordspacing

\bibitem{FRIEDMAN1999275}
\BIBentryALTinterwordspacing
E.~Friedman and H.~Moulin, ``{Three Methods to Share Joint Costs or Surplus},''
  \emph{Journal of Economic Theory}, vol.~87, no.~2, pp. 275 -- 312, 1999.
  [Online]. Available:
  \url{http://www.sciencedirect.com/science/article/pii/S0022053199925346}
\BIBentrySTDinterwordspacing

\bibitem{chu2016large}
T.~Chu, S.~Qu, and J.~Wang, ``{Large-Scale Multi-Agent Reinforcement Learning
  using Image-Based State Representation},'' in \emph{IEEE 55th Conference on
  Decision and Control (CDC)}, 2016, pp. 7592--7597.

\bibitem{shapley1992}
U.~Faigle and W.~Kern, \emph{\BIBforeignlanguage{Undefined}{The Shapley Value
  for Cooperative Games under Precedence Constraints}}, ser. Memorandum.\hskip
  1em plus 0.5em minus 0.4em\relax University of Twente, Faculty of
  Mathematical Sciences, 1992, no. 1025.

\bibitem{vstrumbelj2014explaining}
\BIBentryALTinterwordspacing
E.~{\v{S}}trumbelj and I.~Kononenko, ``{Explaining Prediction Models and
  Individual Predictions with Feature Contributions},'' \emph{Knowledge and
  Information Systems}, vol.~41, no.~3, pp. 647--665, 2014. [Online].
  Available:
  \url{https://moodle.telekom.ftn.uns.ac.rs/pluginfile.php/13342/mod_folder/content/0/Feature%20importance%20paper.pdf?forcedownload=1}
\BIBentrySTDinterwordspacing

\bibitem{tallon2020explainable}
A.~Tall{\'o}n-Ballesteros and C.~Chen, ``{Explainable AI: Using Shapley Value
  to Explain Complex Anomaly Detection ML-Based Systems},'' \emph{Machine
  Learning and Artificial Intelligence: Proceedings of MLIS 2020}, vol. 332, p.
  152, 2020.

\bibitem{lillicrap2015continuous}
T.~Lillicrap, J.~J. Hunt, A.~Pritzel, N.~Heess, T.~Erez, Y.~Tassa, D.~Silver,
  and D.~Wierstra, ``{Continuous Control with Deep Reinforcement Learning},''
  \emph{CoRR}, vol. abs/1509.02971, 2016.

\bibitem{henderson2017deep}
P.~Henderson, R.~Islam, P.~Bachman, J.~Pineau, D.~Precup, and D.~Meger, ``{Deep
  Reinforcement Learning that Matters},'' in \emph{Proceedings of the AAAI
  Conference on Artificial Intelligence}, vol.~32, no.~1, 2018.

\bibitem{SSDOpenSource}
E.~Vinitsky, N.~Jaques, J.~Leibo, A.~Castenada, and E.~Hughes, ``{An Open
  Source Implementation of Sequential Social Dilemma Games},''
  \url{https://github.com/eugenevinitsky/sequential_social_dilemma_games/issues/182},
  2019, gitHub repository.

\bibitem{gini1912variabilita}
C.~Gini, ``{Variabilit{\`a} e Mutabilit{\`a}},'' \emph{Reprinted in Memorie di
  metodologica statistica (Ed. Pizetti E}, 1912.

\bibitem{huber2004robust}
P.~J. Huber, \emph{Robust statistics}.\hskip 1em plus 0.5em minus 0.4em\relax
  John Wiley \& Sons, 2004, vol. 523.

\bibitem{maronna2019robust}
R.~A. Maronna, R.~D. Martin, V.~J. Yohai, and M.~Salibi{\'a}n-Barrera,
  \emph{Robust Statistics: Theory and Methods (with R)}.\hskip 1em plus 0.5em
  minus 0.4em\relax John Wiley \& Sons, 2019.

\bibitem{hampel2011robust}
F.~R. Hampel, E.~M. Ronchetti, P.~J. Rousseeuw, and W.~A. Stahel, \emph{Robust
  Statistics: The Approach based on Influence Functions}.\hskip 1em plus 0.5em
  minus 0.4em\relax John Wiley \& Sons, 2011, vol. 196.

\end{thebibliography}
%


%








\section*{Acknowledgements}
We thank Frédéric Herbreteau, Adrien Bennetot and Léo Heidelberger for their help and support. N. Díaz-Rodríguez is currently supported by the Spanish Government Juan de la Cierva Incorporación contract (IJC2019-039152-I).

{\section{Supplementary Material}
\label{ref:appendix}
Supplementary material can be accessed here: \url{https://bit.ly/3xG7ZXy}} 


\end{document}